\documentclass[10pt,A4pape,conference]{IEEEtran}
\usepackage{cite}
\usepackage{amsmath,amssymb,amsfonts}
\usepackage{algorithmic}
\usepackage{graphicx}
\usepackage{textcomp}
\usepackage{xcolor}
\usepackage{array}
\usepackage{hyperref}
\usepackage{subfigure}

\usepackage{tabularx}
\usepackage{color, colortbl}
\usepackage[misc]{ifsym}
\usepackage{xcolor}
\definecolor{lime}{rgb}{0.88,2,10}

\newcommand*{\Resize}[2]{\resizebox{#1}{!}{$#2$}}%

\renewcommand{\baselinestretch}{0.9993}
\usepackage{subcaption}
\usepackage{wrapfig}
\usepackage[linesnumbered,ruled,vlined]{algorithm2e}
\usepackage{xpatch}
\usepackage{url}
\usepackage{multirow}
\usepackage{multicol}
\usepackage{wrapfig}

\usepackage[export]{adjustbox}
\usepackage{nomencl}
\usepackage{siunitx}
\usepackage{blindtext}
\usepackage[T1]{fontenc}
\usepackage[spaces,hyphens]{xurl}
\usepackage{float}
\usepackage{comment}

\usepackage{diagbox}

\SetAlFnt{\small}

\SetKwComment{Comment}{$\triangleright$\ }{}

\usepackage[utf8]{inputenc}
\usepackage{enumitem}

\SetKwInput{kwInit}{Initialize}

\usepackage{booktabs}

\def\BibTeX{{\rm B\kern-.05em{\sc i\kern-.025em b}\kern-.08em
    T\kern-.1667em\lower.7ex\hbox{E}\kern-.125emX}}

\usepackage[numbers,sort&compress]{natbib}
\usepackage[font={small}]{caption} %it

\newcommand{\fref}[1]{Fig.~\ref{#1}}

\newcommand{\sref}[1]{Section~\ref{#1}}
\usepackage[misc]{ifsym}

\usepackage[outline]{contour}
\contourlength{0.2pt}
\contournumber{15}

\newcommand\HUGE{\fontsize{22.3}{25}\selectfont}

 \usepackage{fancyhdr}
\fancypagestyle{mystyle}{
    \chead{\large The 2024 IEEE International Conference on Big Data (IEEE BigData 2024)}}

\begin{document}

\title{\HUGE DIS-Mine: Instance Segmentation for Disaster-Awareness in Poor-Light Condition in Underground Mines}
%DIS-Mine: Instance Segmentation for Enhancing Disaster Detection and First Responder Safety in Low-Light Underground Mining Environments}

%Disaster Detection in Low-Light Underground Mining: Enhancing First Responder Safety Through Instance Segmentation}
% Conference Paper Title*\\

\author{\IEEEauthorblockN{
Mizanur Rahman Jewel\IEEEauthorrefmark{2}, Mohamed Elmahallawy\IEEEauthorrefmark{3}, Sanjay Madria\IEEEauthorrefmark{2}, Samuel Frimpong\IEEEauthorrefmark{4}}  
      \IEEEauthorblockA{%
 \IEEEauthorrefmark{2}Computer Science Department, Missouri University of Science and Technology, Rolla, MO 65401, USA}
       \IEEEauthorblockA{%
 \IEEEauthorrefmark{3} School of Engineering \& Applied Sciences, Washington State University, Richland, WA 99354, USA}
   \IEEEauthorblockA{%
 \IEEEauthorrefmark{4}Explosive \& Mining
Engineering Department, Missouri University of Science and Technology, Rolla, MO 65401, USA}
%\thanks{$^*$This work was supported by a grant from CDC-NIOSH.} 
Emails:  mj9vc@mst.edu, mohamed.elmahallawy@wsu.edu, madrias@mst.edu, frimpong@mst.edu}

\maketitle
\thispagestyle{mystyle}

\begin{abstract}
Detecting disasters in underground mining, such as explosions and structural damage, has been a persistent challenge over the years. This problem is compounded for first responders, who often have no clear information about the extent or nature of the damage within the mine. The poor-light or even total darkness inside the mines makes rescue efforts incredibly difficult, leading to a tragic loss of life. In this paper, we propose a novel instance segmentation method called DIS-Mine, specifically designed to identify disaster-affected areas within underground mines under low-light or poor visibility conditions, aiding first responders in rescue efforts. DIS-Mine is capable of detecting objects in images, even in complete darkness, by addressing challenges such as high noise, color distortions, and reduced contrast. The key innovations of DIS-Mine are built upon four core components: i) {\em Image brightness improvement}, ii) {\em Instance segmentation with SAM Integration}, iii) {\em Mask R-CNN-Based Segmentation}, and iv) {\em Mask alignment with feature matching}.  On top of that, we have collected real-world images from an experimental underground mine, introducing a new dataset named ImageMine,  specifically gathered in low-visibility conditions. This dataset serves to validate the performance of DIS-Mine in realistic, challenging environments. Our comprehensive experiments on the ImageMine dataset, as well as on various other datasets demonstrate that DIS-Mine achieves a superior F1 score of 86.00\% and mIoU of 72.00\%, outperforming state-of-the-art instance segmentation methods, with at least 15x improvement and up to 80\% higher precision in object detection.

We have made our dataset publicly accessible through {\href{https://drive.google.com/drive/folders/1KVb27J7vU395wa_D8zgtnh-Y205xTiP_?usp=sharingt}{ImageMine Dataset}}.

%that enhances the brightness of dark images to enable effective instance segmentation
%that fuses and aligns the outputs from both the optimized SAM and Mask R-CNN models for superior instance segmentation results.

%that utilizes an optimized version of the Mask R-CNN model on the enhanced images to further enhance segmentation accuracy
%that applies the enhanced images to an optimized Segment Anything Model (SAM) for precise instance segmentation

%Our framework is structured into two stages: First, we employ the "segment anything" concept to extract shared features from a batch of low-visibility images. Next, we use our model to accurately segment objects in the underground mine by leveraging these learned traits. 

%Prior instance segmentation techniques are generally adapted for images with great visibility.
%In this study, we introduce a framework for instance segmentation in underground mine situations with little visibility.
\end{abstract}


\section{Introduction}

Underground mining operations are increasingly hazardous as extraction processes delve deeper into the Earth. As mining extends, increased pressures can lead to significant dangers, such as the ``crushing'' of mine walls and support structures (pillars), or even surface collapse events \cite{debing2016current,zhang2016residual}. In addition, disasters like fires, explosions, or gas and water inundations not only challenge the operational safety of mines but also place miners' lives in jeopardy, requiring immediate, well-coordinated responses \cite{rogers2019automation}. It becomes even more difficult for first responders to enter the mine and assist trapped miners \cite{ashktorab2014tweedr}. Limited information about the type and extent of damage inside the mine can lead to delays and increased casualties \cite{karlsson2020preparedness}. The safety of mine workers depends on various interrelated factors, including real-time environmental awareness, hazard identification, communication, training, and experience, all of which are critical during emergencies \cite{onifade2021towards}. For emergency operations to be effective, responsible individuals/first responders must be prepared to manage and delegate essential tasks.
%While autonomous mining machines offer promising solutions to mitigate these dangers by assisting or replacing humans in risky tasks, some complex mining operations still necessitate human involvement. 
%\red{I added the above paragraph and rewrote the abstract.I will continue working on the paper later today.}
%These segments provide essential context for subsequent high-level tasks, including object recognition, scene understanding, and medical imaging analysis. 

%Deep learning techniques, in particular convolutional neural networks (CNNs), have significantly advanced semantic segmentation. Multiple methods have shown remarkable results on benchmark datasets, including . 
%On the other hand, instance segmentation aims to detect and delineate each distinct object within an image, going beyond semantic limitations. Instance segmentation transcends the task of object class identification by additionally demanding the precise localization of each object instance at the pixel level.

%This work presents a framework to segment the environment in low-light conditions in underground mines accurately.  Underground mining environments are characterized by absolute obscurity as a result of the lack of natural light. Additionally, during emergency scenarios involving potential explosions or structural failures of the walls or roof, the electrical power supply is likely to be severed. 

\begin{figure}[!t]
    \centering
    \includegraphics[width=0.35\textwidth,height=5.5cm]{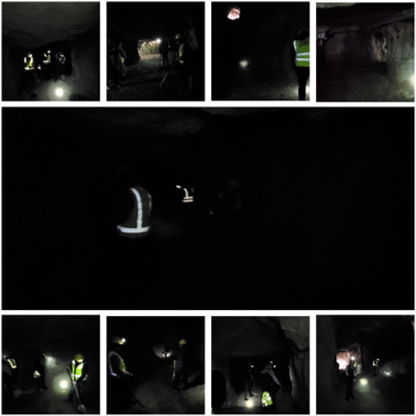}
    \caption{Samples from our underground mine dataset captured under extremely low-light conditions.}
    \label{fig:um_example_image}
\end{figure}
One potential solution is to develop an advanced image segmentation scheme aimed at partitioning an image into coherent regions or segments. This approach could provide first responders with critical information during disasters in mines, helping them identify safe paths and locate miners in need of assistance. Traditional image segmentation methods, such as fully convolutional networks (FCNs) \cite{7298965}, U-Net \cite{ronneberger2015u}, and DeepLab \cite{7913730}, are capable of achieving high accuracy in segmenting images into different categories or classes. However, these methods treat all objects of the same category uniformly, labeling them with the same class, which poses a significant limitation in our context.

This limitation can be addressed by using instance segmentation, which detects and delineates each distinct object within an image, surpassing the semantic segmentation limitations. Instance segmentation not only identifies the class of an object but also precisely localizes each object instance at the pixel level. However, current instance segmentation approaches share a critical drawback:  they perform well {\em only} in segmenting different categories in images captured in ``daylight'' or ``well-lit'' environments but struggle with ``low-visibility'' or ``dark images''. The presence of diminished light levels frequently distorts colors and reduces the contrast between backgrounds and objects. This issue is particularly problematic in underground mining scenarios,{\em especially during disasters, where the environment is often dark, and existing methods fail to detect anything within these conditions.}

Although some initial steps have been taken to address these challenges in instance segmentation studies \cite{chen2023instance,9049390, 10.1145/3343031.3350926}, the results achieved in very dark environments remain suboptimal, with models sometimes achieving less than 50\% F1-score on simple datasets. {\em This leaves a significant open question: is there an instance segmentation approach capable of categorizing objects within images captured in extremely dark environments with acceptable accuracy?}

%These limitations are prevalent in real-world applications, including underground mine navigation, nighttime autonomous driving, low-light surveillance, and various industrial inspection tasks. Instance segmentation performance suffers significantly under low-visibility conditions due to the inherent limitations of image data in such scenarios. In low-visibility environments, segmentation is problematic due to the high level of noise in the images. 

%While doing the rescue operation in those cases it is crucial to get the environmental situation of the whole mine which involves segmenting the roads, walls, people, or necessary objects.  The segmentation results will be useful in assisting the rescue crew in carrying out a successful rescue operation.

{\bf Contribution.}  In response to the aforementioned challenges, this paper proposes a cutting-edge instance segmentation model, named DIS-Mine, capable of accurately identifying objects within images captured under poor-light conditions, including complete darkness. To enhance the model's performance, we also introduce an image enhancement technique specifically designed to improve the quality of dark images, thereby enabling more accurate predictions. To validate the proposed techniques, we have collected a real-world dataset from an experimental mine and manually annotated a subset of images into five classes: road, wall, roof, people, and corridor within the underground mine. Additionally, we develop an automatic annotation pipeline that can be used to label the remaining images, as well as new images from similar environments. In short, we make
the following contributions:
% This pipeline consists of aligning mask output from two models the Mask R-CNN\cite{he2017mask} and SAM where both models are trained on the manually and automated labeled data. We have also used dice loss with focal loss for the mask loss in the Mask R-CNN model. 
\begin{enumerate}[leftmargin=*, label=(\arabic*)]
    \item We propose a novel instance segmentation approach, DIS-Mine, designed to accurately detect and segment each class in images {\em even under poor-light or near-dark conditions}. This approach is applied to the real-world problem of disaster detection in underground mines, assisting first responders in their rescue efforts.

    \item DIS-Mine features four key innovations: i) {\em image brightness improvement component} that enhances the ability to distinguish between foreground and background objects in poor-light images; ii) {\em Instance segmentation with SAM integration component} that applies the enhanced images to an optimized Segment Anything Model (SAM) for precise instance segmentation; iii) {\em Mask R-CNN-based segmentation component} that utilizes an optimized version of the Mask R-CNN model on the enhanced images to further enhance segmentation accuracy; iv) {\em Mask alignment with feature matching} that fuses and aligns the outputs from both the optimized SAM and Mask R-CNN models for superior instance segmentation results. 
    
    %This component enhances image visibility in low-light conditions, allowing DIS-Mine to more effectively detect and segment disaster-affected areas. 
    
    %The integration significantly boosts the accuracy of disaster detection in underground mines, supporting first responders in their rescue efforts.
    
  \item  We collect a real-world image dataset, named {\bf ImageMine}, captured in extremely poor-light conditions from an experimental underground mine (see samples in \fref{fig:um_example_image}). A small subset of these images was manually annotated into six categories: roads, walls, roofs, people, equipment, and corridors—while we also developed an automatic annotation system to label the remaining images.%and any additional images from similar settings.

\item Our experimental evaluation demonstrates that DIS-Mine significantly outperforms state-of-the-art approaches across various datasets, including our collected ImageMine dataset. It achieves an F1 score of 86.00\% and an mIoU of 72.00\%, surpassing existing instance segmentation methods by a substantial margin.
\end{enumerate}
      
   %We have modified the mask loss of Mask R-CNN model to address the challenge .
  
  %and we have also developed a pipeline for automating the labeling process.
  %We have created a pipeline to do instance segmentation of low-visual images considering six labels: road, wall, roof, people, equipment, and corridor. 

  %We are finetuning the Segment Anything model (SAM) \cite{kirillov2023segment} for the automation task.
  
   % \item We have created an image enhancement pipeline.

    %\item 

\section{Related Work}
Instance segmentation task in computer vision aims to identify and delineate each object in an image at the pixel level. Unlike object detection, which provides only bounding boxes around objects, or semantic segmentation, which classifies each pixel into a category without distinguishing between instances of the same category, instance segmentation addresses both localization and classification challenges. Instance segmentation approaches primarily relied on traditional machine learning (ML) techniques. Despite these difficulties, recent efforts have begun to make initial strides in adapting instance segmentation techniques to such challenging conditions.

An approach by Arnab et al. \cite{arnab2017pixel} employed per-pixel unary classifiers combined with conditional random fields to maintain spatial consistency. Although these methods were innovative at the time, they faced challenges with the complexity and variability of real-world images. The advent of deep learning (DL) has since revolutionized instance segmentation, a crucial computer vision task that involves outlining individual objects with pixel-level masks. Convolutional neural networks (CNNs) have become the foundation for many state-of-the-art techniques. Among these, Mask R-CNN, introduced by He et al. \cite{he2017mask}, stands out as one of the most influential models. Mask R-CNN extends the faster R-CNN \cite{ren2015faster} object detection framework by adding an additional branch to predict segmentation masks for each region of interest (ROI). This advancement has significantly enhanced accuracy and established a new benchmark for subsequent research in instance segmentation. Another significant advancement in instance segmentation is the hybrid task cascade (HTC) \cite{chen2019hybridtaskcascadeinstance}. HTC enhances accuracy by integrating object detection and semantic segmentation through a multi-stage architecture. Each stage in HTC refines the predictions from the previous one, resulting in more precise and detailed segmentation masks, especially in complex scenes with overlapping objects.

Real-time instance segmentation has also seen considerable progress. Models such as YOLACT \cite{bolya2019yolac} and INSTA-YOLO \cite{mohamed2021instayolorealtimeinstancesegmentation} have been developed to strike a balance between accuracy and speed. YOLACT decomposes the instance segmentation task into parallel sub-tasks, allowing for faster inference times. INSTA-YOLO builds on the YOLO (You Only Look Once) framework, optimizing it for instance segmentation while maintaining real-time performance. These models are vital for applications requiring immediate feedback, such as autonomous driving and robotics. 

The development of large-scale datasets has been crucial for advancing instance segmentation. Datasets like COCO (Common Objects in Context) \cite{lin2015microsoftcococommonobjects} and Cityscapes \cite{cordts2016cityscapes} provide diverse and challenging benchmarks for evaluating model performance. They encompass a wide variety of objects and scenes, enabling researchers to develop and test models under various conditions. Recent research has also explored transformer-based architectures for instance segmentation. The detection transformer (DETR) \cite{DBLP} leverages transformers to model relationships between objects in an image, offering a unified approach to panoptic segmentation. This method has demonstrated promising results, particularly in handling complex scenes with multiple interacting objects.

Despite significant advancements in instance segmentation, challenges remain, such as handling occlusion, varying object scales, and high computational demands. Most importantly, instance segmentation models often perform poorly under low-light conditions, where reduced visibility and increased noise hinder their accuracy. Traditional methods often struggle in such scenarios due to reduced visibility and increased noise, which introduce high-frequency disturbances to neural network feature maps and significantly degrade accuracy. Although some recent works \cite{chen2023instance, Lin2024FeatureDF} have introduced techniques to enhance accuracy by addressing noise in low-light images, the research into instance segmentation in dark or low-illumination environments remains limited. This challenge is especially pronounced in extreme low-light environments like underground mining, where poor visibility severely complicate object detection. {\em In this paper, we propose an innovative solution to these challenges, validated on our dataset collected from extremely dark conditions in an underground mine.}

\section{ImageMine Dataset Construction}\label{sec:ImageMine}
\begin{figure}[!t]
    \centering
    \includegraphics[width=0.5\textwidth]{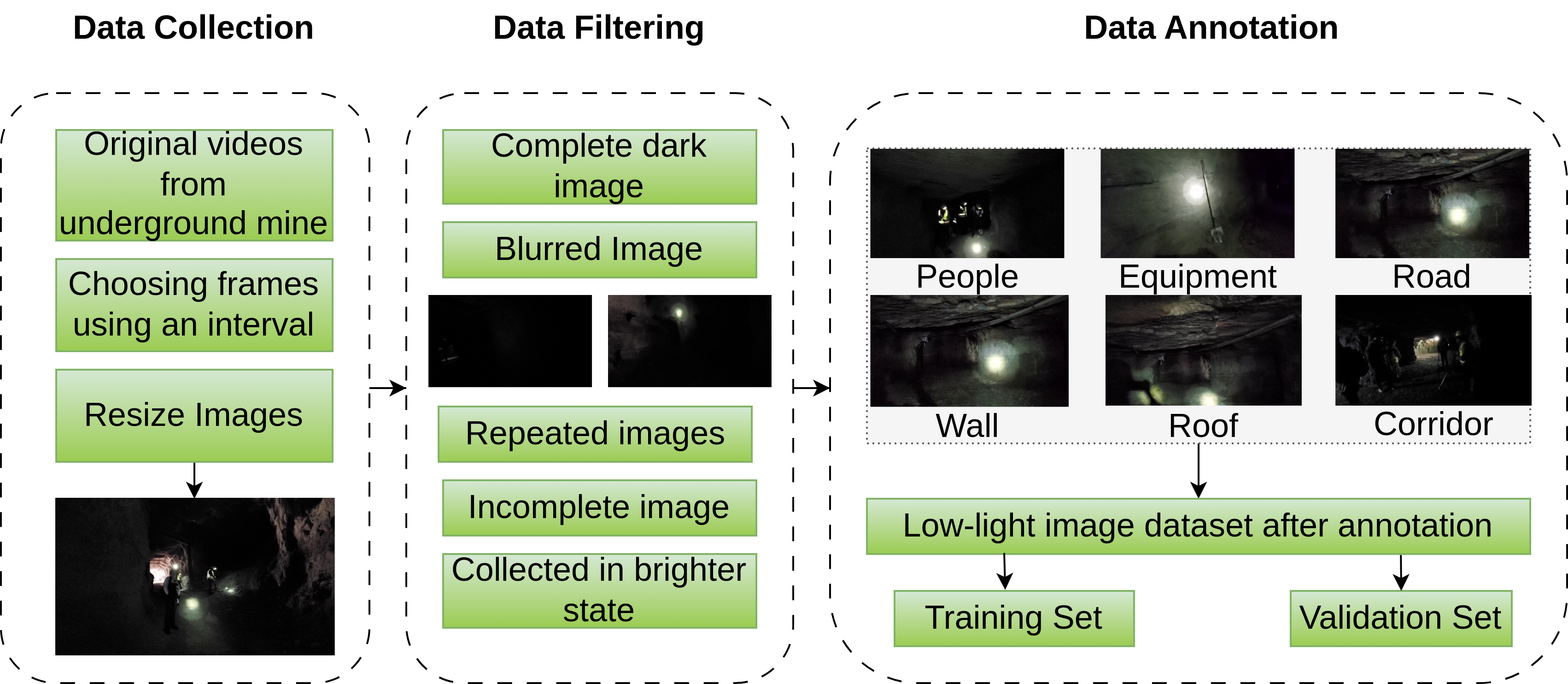}
    \caption{Construction process for our ImageMine dataset. }
    \label{fig:Dataset_pipe}
\end{figure}

This section details the construction of our underground mining image dataset, ImageMine, which is divided into three main phases: Data Collection, Data Filtering, and Data Annotation. \fref{fig:Dataset_pipe} provides an illustration, with each phase explained in detail in the subsequent subsections.

\subsection{Data collection} The underground surveillance videos were collected from an experimental mine located in Rolla, Missouri, USA, {\em under extremely low-light conditions}. From each video, about 100-80 images were extracted from the initial frames for manual annotation, focusing on different target objects. To ensure privacy, all data collected complied with consent regulations, ensuring that no facial information or identifiable features of individual miners were captured.

The image acquisition system utilized a high-resolution surveillance camera, featuring a 12 MP wide-angle lens with a variable aperture of f/1.5 to f/2.4 and a 26mm focal length. The camera provided a 77$^\circ$ field of view and captured images at a maximum resolution of 12 megapixels, supporting video recording in up to 4K ultra-high definition (UHD) at 60 frames per second (FPS), using formats like MP4 and HEVC. This advanced system enabled high-quality image capture even under the poor lighting conditions in underground mines.

The collected ImageMine dataset was annotated into {\bf six} categories: road, wall, roof, people, equipment, and corridor. These labels are selected based on the importance of the situation while conducting a rescue operation in an underground mine. Due to poor lighting, some frames lacked target objects such as equipment, people, or corridors, and these frames were excluded. The remaining annotated images were sorted by category to form the final ImageMine dataset.

\begin{figure*}[!t]
    \centering
    \includegraphics[width=1\linewidth]{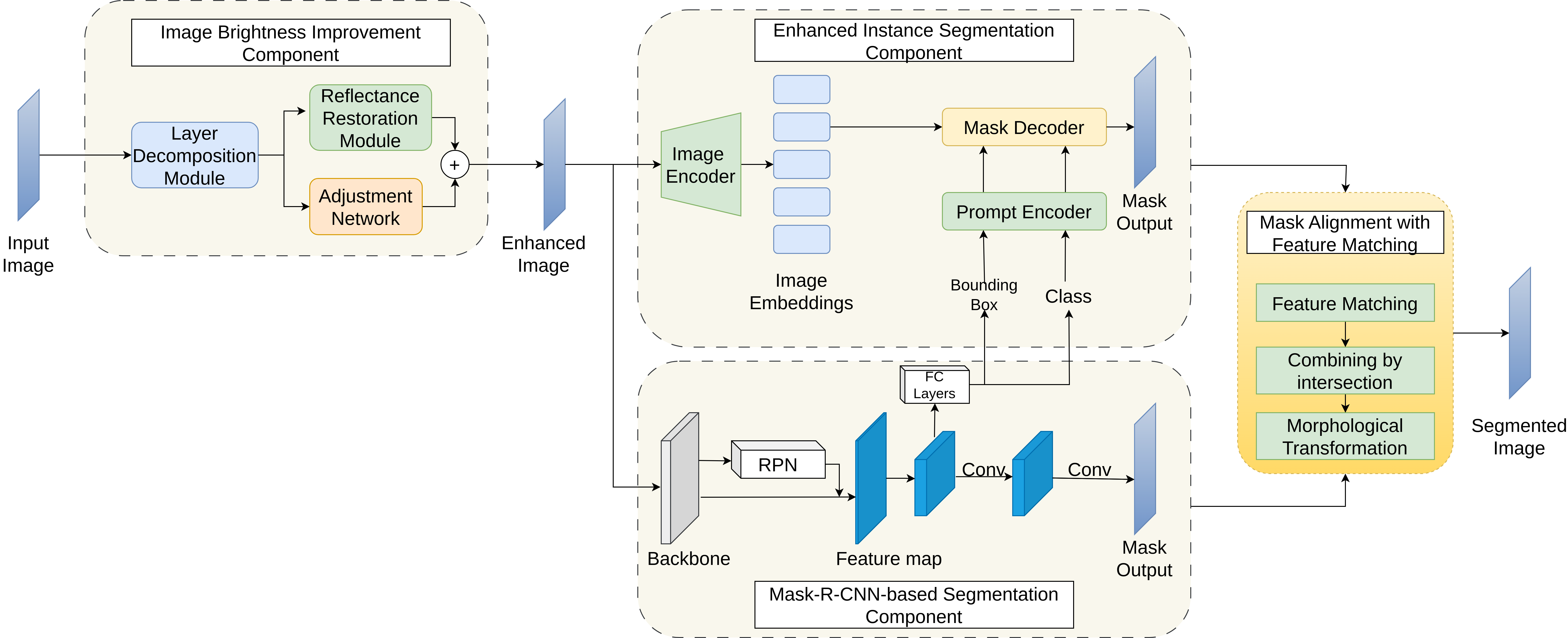}
    \caption{An overview of proposed DIS-Mine instance segmentation framework.}
    \label{fig:segmentation-pipeline}
\end{figure*}

\subsection{Data Filtering}

The original image source of the ImageMine dataset is screened to ensure high quality. This dataset primarily includes images of road, wall, roof, people, equipment, and corridor. However, some images may lack targets, have incomplete targets, or be of poor quality, making identification challenging. For example, some images may appear excessively dark, providing little visibility. Consequently, images with abnormal data must be eliminated manually or automatically during the dataset production process. To address this, we apply an automatic filtering system to efficiently process and remove invalid images, thereby enhancing dataset reproducibility and facilitating collaboration among researchers. Abnormal images that require processing typically include those where severe environmental factors hinder the identification of people and equipment, images that capture only local features due to limited camera views or occlusion, repeated images, and severely dark images collected after mining operations have ceased. Additionally, blurred images from fast-moving targets during video-to-image conversion and images where distant targets are indistinguishable from other equipment due to environmental conditions and distance from the camera also fall into this category.

\subsection{Data annotation}\label{sub:annotation} 
Finally, to label the filtered images, we employed the pretrained segment anything model (SAM) for automatic annotation tasks \cite{kirillov2023segment}. SAM is capable of generating high-quality object masks from input prompts such as points or boxes, making it versatile for various segmentation applications. Initially, SAM was fine-tuned on a limited number of manually annotated datasets of 510 images. Manual annotation was done using VGG image annotator\cite{dutta2016via}. Once fine-tuned, it was applied for automatic annotation across the remainder of the dataset. Additionally, we selected brighter images to create a subset for training an image enhancement component. To simulate low-light conditions, we introduced noise to this brighter subset by applying Gaussian noise, reducing brightness, and increasing contrast, thus generating low-light pairs of those images.

\section{Proposed Methodology}

This section provides an overview of the DIS-Mine framework (\fref{fig:segmentation-pipeline}), which comprises three core components: Image Enhancement, Instance Segmentation (via the Segment Anything Model - SAM), and Mask R-CNN . The process flow of DIS-Mine is as follows:

\begin{itemize}[leftmargin=*]
    \item Each input image is processed through our proposed image brightness improvement network (\sref{sec:enhance}), which is based on the KinD (Kindling the Darkness) network \cite{10.1145/3343031.3350926}. This network improves image quality by enhancing brightness, contrast, and texture. It works by decomposing, adjusting, and reconstructing the image for optimal enhancement. 

   \item  The improved image is used as input for the instance segmentation with SAM integration component (\sref{sec:inst_seg}), which incorporates SAM as its core. The image is fed into the image encoder, while the bounding box and class prediction are passed into the prompt encoder to generate the instance mask.

    \item Concurrently, the improved image is also fed into our Mask R-CNN-based segmentation component (\sref{sec:Mask-RCNN}), which extends Faster R-CNN to generate bounding boxes, class predictions, and instance masks for each detected object.

    \item  Finally, the outputs from both the instance segmentation with SAM integration component and the Mask R-CNN-based segmentation component are aligned through a series of operations, including feature matching, intersection-based combination, and morphological transformations, which are explained in detail in \sref{sec:feature}.
    
\end{itemize}

\subsection{Image Brightness Improvement Component}\label{sec:enhance}

The image brightness improvement component is developed to enhance the lighting of the ImageMine dataset, aiming for better prediction accuracy. Specifically, we integrate the KinD network \cite{10.1145/3343031.3350926} to the DIS-Mine network, which effectively enhances low-light images by improving visibility and quality in poorly lit environments. Grounded in Retinex theory \cite{McCann2016}, the KinD network decomposes images into reflectance and illumination components. The integrated network consists of three main modules: {\em the layer decomposition module} that separates the input image into reflectance and illumination layers; {\em the reflectance restoration module} that enhances the reflectance layer to recover true colors and details while minimizing noise; and {\em the adjustment network} that modifies the illumination layer to improve overall brightness and contrast. This image enhancement technique is applied to the input images before feeding them into the SAM model or Mask R-CNN model for training.

To begin with, we train this network on a synthetic normal/low-light dataset derived from relatively brighter images in our ImageMine dataset as we discussed in \sref{sub:annotation}. The network architecture includes three functional modules: i) layer decomposition, ii) reflectance restoration, and iii) illumination adjustment net. The normal/low-light pair images first pass through the layer decomposition block, which has two paths: one creates reflectance maps, while the other generates illumination maps. The reflectance path consists of five convolutional layers followed by a sigmoid layer. The first two convolutional layers are for downsampling, the next two are for upsampling, and the final convolutional layer is processed through the sigmoid layer to produce the reflectance map. In contrast, the illumination path consists of three convolutional layers and a sigmoid layer, utilizing features from the reflectance maps.

\begin{algorithm}[!t]
    \SetAlgoLined
    \caption{\footnotesize Training  Process of Image Brightness Component}
    \label{algo:algo_image_enhancement_training}
    \KwIn{Low-Light Image: $I_{\text{low}}$ , Normal Image: $I_{\text{normal}}$}
    \KwOut{Enhanced Image: $I_{\text{EN}}$}
    
    $RM_{\text{low}}, RM_{\text{normal}}, IM_{\text{low}}, IM_{\text{normal}} \leftarrow \text{LDM}(I_{\text{low}}, I_{\text{normal}})$\\
    \Comment{Extract Reflection Map (RM) and Illumination Map (IM) for low and normal images}
    \Comment{LDM indicates the layer decomposition module}
    
    Calculate loss between $RM_{\text{low}}, RM_{\text{normal}}$\\
    Calculate loss between $IM_{\text{low}}, IM_{\text{normal}}$\\
    
    $RM_{\text{restored}} \leftarrow \text{Reflectance Restoration }(RM_{\text{low}}, IM_{\text{low}})$\\
    \Comment{Restore reflection map based on low-light image}
    
    $IM_{\text{adjusted}} \leftarrow \text{Illumination Adjustment}(IM_{\text{low}})$\\
    \Comment{Adjust illumination map for better contrast}

    Calculate loss between $RM_{\text{restored}}, RM_{\text{normal}}$\\
    Calculate loss between $IM_{\text{adjusted}}, IM_{\text{normal}}$\\

    Minimize the loss and update weights\\
    $I_{\text{EN}} \leftarrow \text{Combine}(RM_{\text{restored}}, IM_{\text{adjusted}})$\\
    \textbf{Return} $I_{\text{EN}}$
\end{algorithm}

Following layer decomposition, the reflectance maps proceed to the reflectance restoration module, which includes a 5-layer U-Net \cite{ronneberger2015u}, a convolutional layer, and a sigmoid layer. Simultaneously, the illumination maps go through an illumination adjustment layer, comprising two convolutional and rectified linear unit (ReLU) layers, and a convolutional layer followed by a sigmoid layer. The outputs from both the reflectance restoration and illumination adjustment modules are combined pixel-wise to produce the final enhanced image, which increases light in relatively dark regions while maintaining brightness in lighter areas. 

Algorithm~\ref{algo:algo_image_enhancement_training}  offers a explanation of the entire process.

\subsection{ Instance Segmentation with SAM Integration Component}\label{sec:inst_seg}
We propose an enhanced version of the segment anything model (SAM) \cite{kirillov2023segment}. SAM is a promptable segmentation system that enables users to segment objects within an image using minimal input, such as a single click. It is designed to generalize to unfamiliar objects and images without the need for additional training. Specifically, SAM comprises three primary encoders: the image encoder, the prompt encoder, and the mask decoder.

\begin{figure}[!t]
    \centering
    \includegraphics[width=0.5\textwidth]{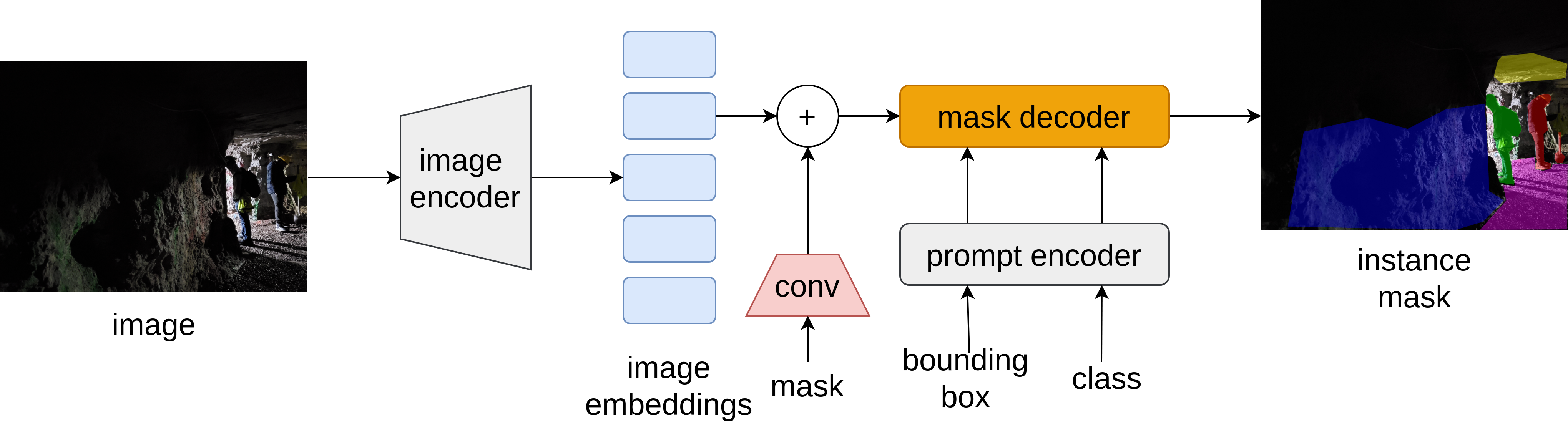}
    \caption{Illustration of the instance segmentation with SAM integration component.}
    \label{fig:SAM}
\end{figure}

The image encoder processes the input image to create an embedding that extracts relevant features, specifically adapted for high-resolution images. The prompt encoder accepts various types of prompts, distinguishing between sparse prompts (points, boxes, and texts) and dense prompts (masks). Finally, the mask decoder generates the segmentation mask based on the encoded image and prompts, processing the information through a down-sampling convolutional layer and concatenating it with the image embeddings. SAM has been trained on a massive dataset of 11 million images and 1.1 billion masks, enabling it to achieve strong zero-shot performance across a wide range of segmentation tasks. For our enhanced SAM version, we utilize only sparse prompts, which consist of boundary boxes and class labels predicted by the Mask R-CNN model (see \fref{fig:SAM}). In Algorithm~\ref{algo:algo_sam}, we outline the process of our instance segmentation component.

\begin{algorithm}[!t]
    \SetAlgoLined
    \caption{\fontsize{9.7pt}{11pt}\selectfont  Training process of Instance Segmentation }
    \label{algo:algo_sam}
    \KwIn{Enhanced Image: $I_{\text{EN}}$, Class Prediction: $Class$, Bounding Box: $BB$}
    \KwOut{$Mask$}
    Image input for SAM, $I_{\text{i}} \leftarrow I_{\text{EN}}$\\
    Prompt input for SAM, $P_{\text{i}} \leftarrow Class, BB$\\
    $Mask \leftarrow  \text{SAM}(I_{\text{i}}, P_{\text{i}})$\\
    \textbf{Return} $Mask$
\end{algorithm}

\subsection{Mask R-CNN-based Segmentation Component}\label{sec:Mask-RCNN}

In this component, we extend the functionality of Mask R-CNN \cite{he2017mask}, which builds upon Faster R-CNN \cite{ren2015faster} by introducing an additional branch for predicting segmentation masks on each region of interest (RoI). This branch works alongside the existing branches for classification and bounding box regression, allowing the model to not only detect objects but also generate high-quality pixel-level masks for each detected instance.

The Mask R-CNN framework features two key innovations: RoIAlign and the mask branch. RoIAlign improves upon the RoIPool method used in Faster R-CNN by avoiding the quantization of RoI boundaries, ensuring precise spatial alignment, and preserving finer details. The mask branch is composed of a small fully convolutional network (FCN) that generates a binary mask for each RoI. Crucially, this mask branch operates independently from the class prediction, enabling more accurate and refined segmentation.

The overall loss of the Mask R-CNN model is divided into three parts. The first two losses represent the Fast R-CNN components: the classification loss and the bounding box regression loss, which can be expressed as follows:
\begin{equation}\label{eq_mask}
    \mathcal{L}_{\text{Fast R-CNN}} = \ell_{\text{class}} + \ell_{\text{box}}
\end{equation}
where $\ell_{\text{class}}$ is the classification loss, which computes the difference between the predicted class probabilities and the ground truth labels using a softmax cross-entropy loss. $\ell_{\text{box}}$ is the bounding box regression loss, responsible for minimizing the difference between the predicted bounding box coordinates and the ground truth, typically using a smooth $\ell_1$ loss. We can express \eqref{eq_mask} in a more generalized form as follows:
\begin{equation}
   \Resize{8cm}{ \ell(p_i, t_i) = \frac{1}{N_{\text{class}}}\sum_{i}\ell_{\text{class}}(p_i,p_i^*) + \lambda\frac{1}{N_{\text{box}}}\sum_{i}p_i^*\ell_{\text{box}}(t_i,t_i^*)}
\end{equation}
where $N_{\text{class}}$ and $N_{\text{box}}$ represent the number of ROIs used for classification and bounding box regression, respectively. The terms $p_i$ and $p_i^*$ correspond to the predicted and ground truth labels for the $i$-th ROI in the classification task, while $t_i$ and $t_i^*$ denote the predicted and ground truth bounding box parameters for the $i$-th ROI.

\begin{algorithm}[!t]
    \SetAlgoLined
    \caption{\fontsize{8.9pt}{11pt}\selectfont Training process of Mask R-CNN  Component}
    \label{algo:algo_Mask}
    \KwIn{Enhanced Image: $I_{\text{EN}}$}
    \KwOut{Instance Mask: $Mask$, Class Prediction: $Class$, Bounding Box: $BB$}

    Initialize Mask R-CNN model\\
    Add modified mask loss to the multitask loss\\
    Minimize the loss and update weights\\
    $Mask, Class, BB \leftarrow  \text{Mask R-CNN}(I_{\text{EN}})$\\

    \textbf{Return} $Mask, Class, BB$

\end{algorithm}

The third loss term, $\ell_{\text{mask}}$, is unique to Mask R-CNN and enables the model to perform accurate pixel-wise segmentation in addition to object detection. This term allows Mask R-CNN to extend the capabilities of Faster R-CNN by not only identifying and locating objects but also generating precise segmentation masks for each instance. By introducing this mask loss, Mask R-CNN becomes a powerful tool, for instance, segmentation, transforming the traditional object detection framework into one capable of detailed pixel-level object delineation.

In our enhanced version of Mask R-CNN, we replace the standard binary cross-entropy mask loss $\ell_{\text{mask}}$ with a combination of weighted dice loss and focal loss. This modification is aimed at improving segmentation precision, particularly in cases of class imbalance or complex object boundaries. The new mask loss, combining these two functions, further enhances the model's instance segmentation performance. The modified loss can be expressed as:
\begin{equation}
   \ell_{\text{mask}} = 
   \ell_{w\text{-Dice}} + \ell_{\text{Focal}}
\end{equation}
The Dice loss optimizes mask quality by ensuring overlap between predicted and ground truth masks. However, in cases of class imbalance, the weighted Dice loss, $\ell_{w\text{-Dice}}$, further improves segmentation by assigning a higher weight to the foreground class. This modification ensures that smaller or less frequent objects are segmented with greater accuracy. The weighted Dice loss, $\ell_{w\text{-Dice}}$,  can be expressed as:
\begin{equation}
    \ell_{\text{W-Dice}} = 1 - \frac{\sum_{c} w_c \cdot \left( 2 \cdot \text{TP}_c \right)}{\sum_{c} \left( w_c \cdot (\text{TP}_c + \text{FP}_c + \text{FN}_c) \right)}
\end{equation}
where  \( w_c \) is the weight for a class \(c\), \( \text{TP}_c \) is the true positive count for class \( c \), \( \text{FP}_c \) is the false positive count for class \( c \), and \( \text{FN}_c \) is the false negative count for class \( c \).

The focal loss $\ell_{\text{focal}}$, reduces the influence of easy-to-learn examples, shifting focus towards challenging pixels, which are often prevalent in low-light or complex environments. This is particularly useful for improving segmentation performance in darker images similar to our dataset, ImageMine, where difficult examples dominate. The focal loss \( \ell_{\text{Focal}} \) for binary classification is defined as:
\begin{equation}
    \ell_{\text{Focal}} = - \alpha \cdot (1 - \hat{p})^\gamma \cdot \log(\hat{p}) - (1 - \alpha) \cdot \hat{p}^\gamma \cdot \log(1 - \hat{p})
\end{equation}
where \( \hat{p} \) is the predicted probability for the positive class, \( \alpha \) is a weighting factor for the class, \( \gamma \) is the focusing parameter.

Thus, the final loss function utilized by our enhanced Mask R-CNN is a multi-task objective that integrates three components: classification loss, bounding box regression loss, and mask prediction loss. This comprehensive loss function can be expressed as:

\begin{equation}
    \mathcal{L}_{\text{total}} = \underbrace{\ell_{\text{class}} + \ell_{\text{box}}}_{\ell_\text{Fast-R-CNN}} + \underbrace{\ell_{w\text{-Dice}} + \ell_{\text{focal}}}_{\ell_{\text{enhanced-mask}}}
\end{equation}

A detailed explanation of the entire process has been showed in Algorithm~\ref{algo:algo_Mask}.

\subsection{Mask Alignment with Feature Matching}\label{sec:feature}

This component is designed to select the most effective mask outputs from both segmentation models. Initially, the masks generated by the instance segmentation with SAM integration component and the Mask R-CNN-based segmentation model undergo an alignment process using a feature-matching algorithm, specifically the oriented fAST and rotated brief (ORB) method \cite{6126544}. After alignment, we combine the masks by taking their intersection, highlighting the common areas identified by both models.

Next, we refine the final instance mask output using morphological operations that combine dilation and erosion. Additionally, we implement rule-based operations based on the mine's structure. We divide the image into 4x4 grids and classify instances according to their positions within these grids. For example, any object detected in the lower row of the grid is designated as part of the road class, even if it extends into the upper grid. The complete process of this component is detailed in Algorithm~\ref{algo:mask_Align}.

\begin{algorithm}[!t]
    \SetAlgoLined
    \caption{Mask Alignment Process}
    \label{algo:mask_Align}
    \KwIn{Enhanced Image: $I_{\text{EN}}$}
    \KwOut{Refined-Mask: $Mask_{\text{final}}$}
    $Mask_{\text{2}}, Class, BB \leftarrow  \text{Mask R-CNN-based Segmentation Component}(I_{\text{EN}})$\\
    \fontsize{8.7pt}{11pt}\selectfont  {$Mask_{\text{1}} \leftarrow \text{ Instance Segmentation component}(I_{{i}}, Class, BB)$}\\
    Alignment using ORB feature-matching algorithm\\
    $aligned\_mask_{1} \leftarrow \text{ORB\_alignment}(Mask_{\text{1}}, Mask_{\text{2}})$\\
    $aligned\_mask_{2} \leftarrow \text{ORB\_alignment}(Mask_{\text{2}}, Mask_{\text{1}})$\\
    $combined\_mask \leftarrow \text{intersection}(aligned\_mask_{1}, aligned\_mask_{2})$\\
    $dilated\_mask \leftarrow \text{dilation}(combined\_mask)$\\
    $Mask_{\text{final}} \leftarrow \text{erosion}(dilated\_mask)$\\
    
    \textbf{Return} $Mask_{\text{final}}$
\end{algorithm}

%\begin{figure*}[h]
 %   \centering
  %  \includegraphics[width=\linewidth]{images/icdm-Segment Anything.png}
   % \caption{Mask Generation using Segment Anything Model}
    %\label{fig:mask_gen_sam}
%\end{figure*}

\section{Experiments and Results}

\subsection{Experimental setup}

% {\bf Datasets.} To evaluate DIS-Mine against state-of-the-art approaches, we consider various benchmark datasets for poor-light conditions segmentation tasks, including the LIS dataset \cite{chen2023instance}, DsLMF+ \cite{zhang2024dslmf}, and DarkVisionNet \cite{jin2023darkvisionnet}, along with our collected dataset, ImageMine (for more information about this dataset, please refer to \sref{sec:ImageMine}). Below is a brief description of each dataset:
{\bf Datasets.} To evaluate DIS-Mine against state-of-the-art approaches, we consider a couple of benchmark datasets for poor-light conditions segmentation tasks, including the LIS dataset \cite{chen2023instance}, DsLMF+ \cite{zhang2024dslmf}, along with our collected dataset, ImageMine (for more information about this dataset, please refer to \sref{sec:ImageMine}). Below is a brief description of other datasets:
\begin{itemize}
    \item {\bf LIS dataset\cite{chen2023instance}.} This dataset comprises 2,230 pairs of low-light and normal-light images collected from diverse indoor and outdoor scenes. It aims to tackle the challenges of instance segmentation in extremely low-light conditions, where traditional methods often struggle. In our experiments, we specifically focused on the compelling low-light images from this dataset.

    \item {\bf DsLMF+\cite{zhang2024dslmf}.} This dataset consists of 138,004 images captured in underground longwall mining faces, encompassing six classes: mine personnel, hydraulic support guard plates, large coal, towlines, miners’ behaviors, and mine safety helmets.

    % \item {\bf DarkVisionNet\cite{jin2023darkvisionnet}.} This benchmark dataset is specifically designed for low-light imaging tasks, consisting of aligned RGB-NIR (Near-Infrared) image pairs that capture the same scenes under low-light conditions. It includes 900 static scenes featuring objects from 15 different categories, as well as 32 dynamic scenes with objects from 4 categories.
\end{itemize}

\begin{table}[!t]
\centering
\setlength{\tabcolsep}{1.2em}  % Increase the space between columns
\renewcommand{\arraystretch}{1.5}  % Increase the space between rows
\caption{Simulation Parameters}
\label{tab:hyperparameters}
\resizebox{\columnwidth}{!}{ 
\begin{tabular}{p{1.9cm}|p{2.35cm}|p{2.cm}|p{2.35cm}} 
\toprule
 \multirow{2}{*}{\textbf{Hyperparameter}} & \multicolumn{3}{c}{\textbf{DIS-Mine Components}} \\ \cline{2-4}
& {Image Improvement} & {Optimized SAM} & {Mask R-CNN-based} \\
\midrule\midrule
{Learning Rate} & 0.001 & 0.0001 & 0.001 \\ \hline
{Batch Size} & 16 & 32 & 4 \\ \hline
{Optimizer} & Adam & Adam & SGD \\ \hline
{Epochs} & 100 & 80 & 80 \\ \hline
{Loss Function} & MSE & Cross-Entropy & Multi-task Loss \\ \hline
{Backbone} & - & ViT & ResNet101 \\ \hline
{Image Size} & 256x256 & 1024x1024 & 1024x1024 \\ \hline
{Augmentation} & Random Crop, Flip & Random Crop & Random Crop \\ \hline
{Regularization} & L2 Regularization & Dropout & L2 Regularization \\ 
\bottomrule
%{Normalization} & Batch Normalization &{\footnotesize Layer Normalization} & {\footnotesize Batch Normalization} \\ \hline
\end{tabular}
} 
\end{table}

{\bf Training Procedure.} The training procedure of DIS-Mine encompasses the training of each individual component. The hyperparameters used for training each component are summarized in Table~\ref{tab:hyperparameters}. For the automatic annotation task, we employed the optimized SAM model, training it with pre-trained model weights on manually annotated images. Masks for each image were generated using the optimized SAM model, and these image-mask pairs were subsequently utilized to train our third component, the Mask R-CNN model. Additionally, the SAM model was retrained on both manually and automatically annotated datasets.

{\bf Environment.} We conducted our training on a DELL R740xa equipped with 238 GB of RAM, an Nvidia A100 GPU, and 80 GB of GPU memory. Various versions of Python 3 and packages from the PyTorch framework were utilized.

\subsection{Baselines}

We compare DIS-Mine with several state-of-the-art instance segmentation methods, including SAM \cite{kirillov2023segment}, Mask R-CNN \cite{he2017mask}, Mask2Former \cite{cheng2021mask2former}, and Instance Segmentation in Dark (ISD) \cite{chen2023instance}. These models serve as strong baselines in the field of semantic segmentation, particularly under poor lighting conditions, allowing for a comprehensive assessment of the effectiveness of DIS-Mine. Below is a brief description of each model:
\begin{itemize}
    \item {\bf SAM \cite{kirillov2023segment}.} SAM is a versatile prompt-based segmentation model capable of generating instance masks across various contexts.

    \item {\bf Mask R-CNN \cite{he2017mask}.} Mask R-CNN is known for its robust performance in object detection and instance segmentation, extending Faster R-CNN with a pixel-level mask prediction branch.

    \item {\bf Mask2Former \cite{cheng2021mask2former}.} Mask2Former is a universal segmentation model for instance, semantic, and panoptic tasks, using masked attention to improve segmentation accuracy.

    \item {\bf ISD \cite{chen2023instance}.} ISD is designed for low-light conditions, employing adaptive downsampling and disturbance-suppressing learning to reduce noise.
\end{itemize}

% \begin{figure*}[!t]
%     \centering
%     \includegraphics[width=1\linewidth]{images/performance.png}
%     \caption{Performance comparison of different models on ImageMine. %\red{legened is very tiny can you please increase the fonts or add these as subfigures?}
%     }
%     \label{fig:learning-comparison}
% \end{figure*}

{\bf Evaluation metrics.}  We assess the performance of DIS-Mine against our baselines using two widely adopted evaluation metrics in instance segmentation: F1-score and mean intersection over union (mIoU). We choose these metrics due to their effectiveness in measuring both segmentation accuracy and overlap between predicted and ground truth masks. Below is a brief description of each metric: 

\begin{itemize}
\item {\bf F1-Score:} The F1-Score is the harmonic mean of precision and recall, providing a balanced metric that accounts for both false positives and false negatives. In instance segmentation, precision measures how accurately the predicted instance masks align with ground truth masks, while recall assesses how effectively the model detects all relevant instances. A high F1-Score indicates that the model achieves both high precision (few false positives) and high recall (few false negatives), reflecting strong overall performance in detecting and segmenting objects. The mathematical formula for the F1-Score is:

\begin{equation}
    \text {F1-Score} = 2 \times \frac{\text{Precision} \times \text{Recall}}{\text{Precision} + \text{Recall}}
\end{equation}

\item  {\bf IoU \& mIoU.} is a measure of overlap between the predicted mask and the ground truth mask. It is defined as the area of overlap between the predicted and actual instance masks divided by the area of their union. IoU provides a direct assessment of how accurately the model predicts the shape and boundaries of instances. A higher IoU score reflects better segmentation quality, as it indicates more accurate pixel-level alignment between prediction and ground truth. mIoU (mean IoU) refers to the average of IoU values over all classes in a dataset. It measures the overlap between the predicted segmented mask and ground truth for each class which summarizes the model's segmentation accuracy. 
The mathematical formula for the mIoU is:
\begin{equation}
\text{mIoU} = \frac{1}{N} \sum_{i=1}^{N} \frac{|A_i \cap B_i|}{|A_i \cup B_i|}
\end{equation}
where N is the total number of classes, $A_i$ and $B_i$ represent the predicted and ground truth areas of class $i$. $|A_i \cap B_i|$ is the area of overlap between the predicted segmentation $A_i$ and the ground truth $B_i$ for class $i$. $|A_i \cup B_i|$ is the total area covered by both the predicted segmentation and the ground truth for class $i$.

\end{itemize}

\begin{figure*}[!t]
    \centering
    \subfigure[F1-score.]{
        \includegraphics[width=0.315\linewidth]{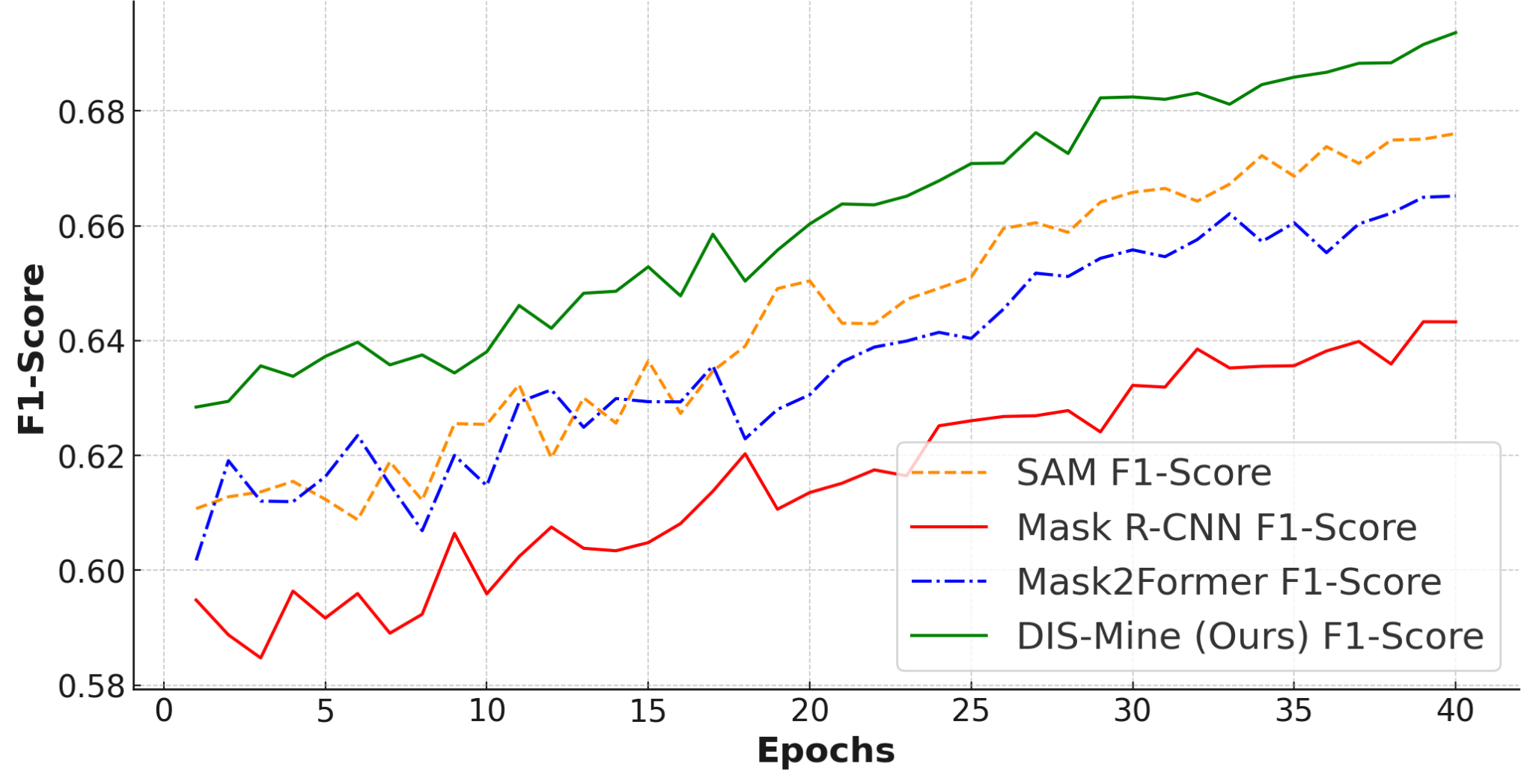}
        \label{fig:f1-comparison-result}
    }
    \subfigure[mIoU score.]{
        \includegraphics[width=0.315\linewidth]{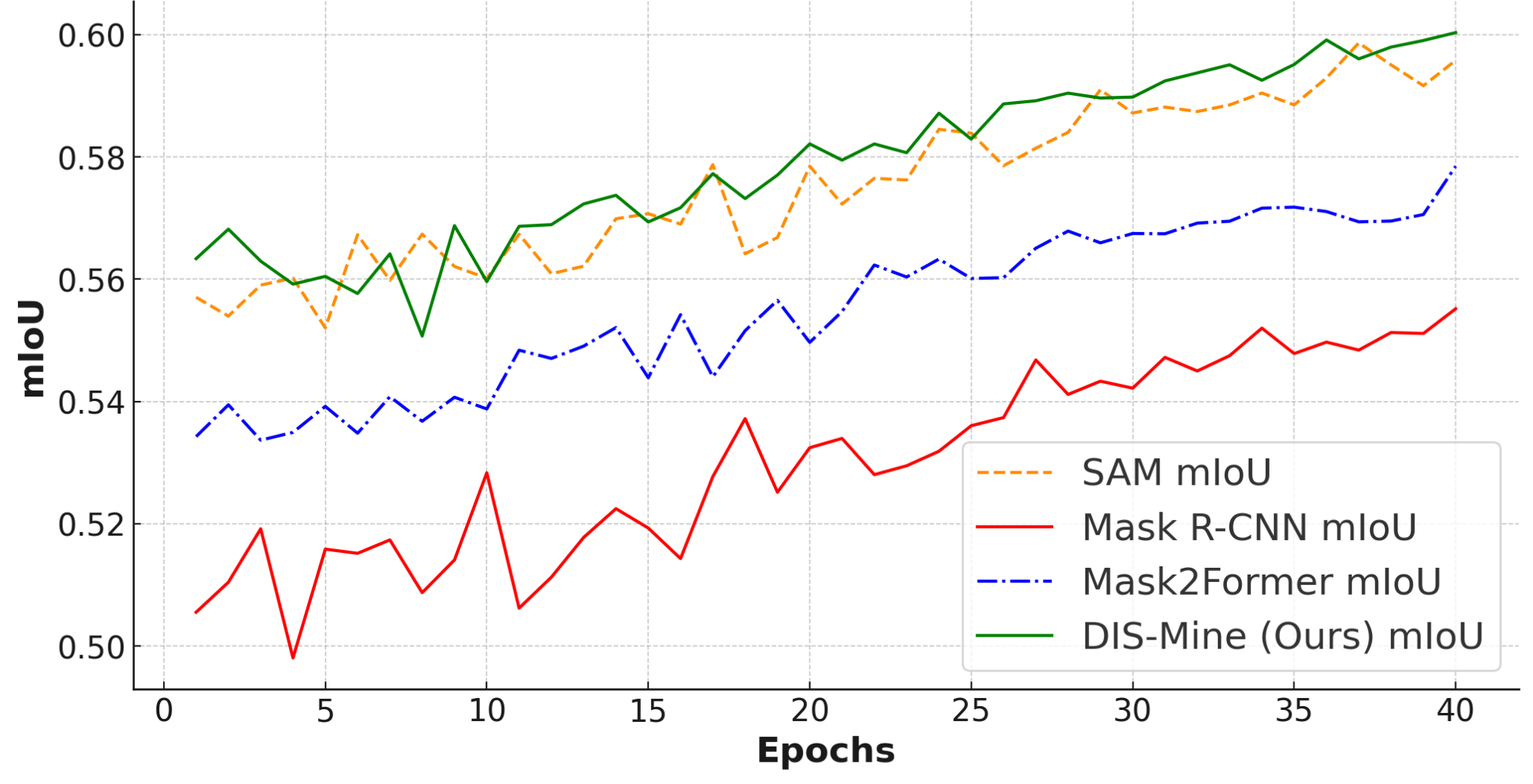}
        \label{fig:mIou-comparison-result}
    }
    \subfigure[Loss-Curve.]{
        \includegraphics[width=0.315\linewidth]{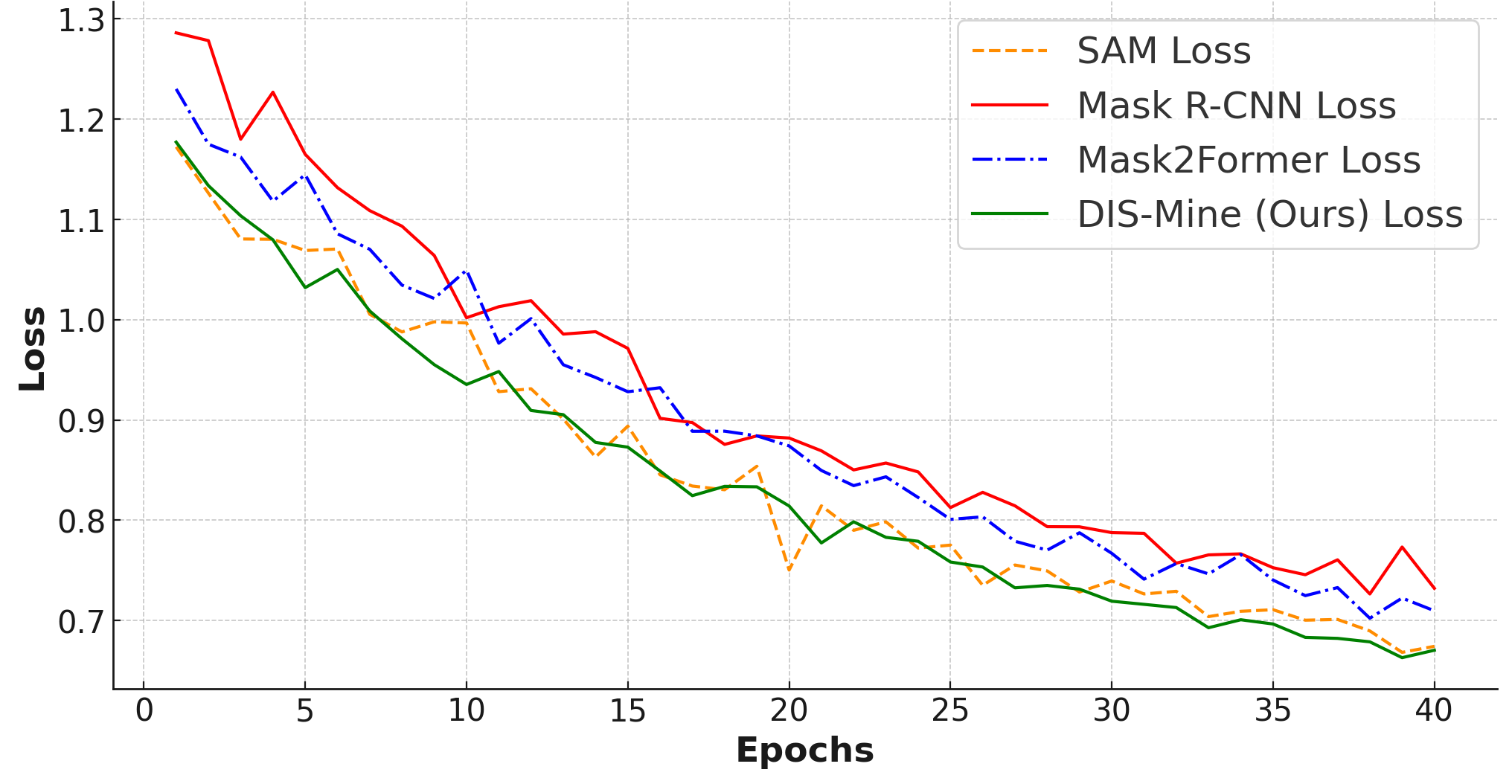}
        \label{fig:loss-comparison}
    }
\caption{Evaluation of DIS-Mine against baselines on various metrics on the ImageMine dataset.}
    \label{fig:learning-comparison}
\end{figure*}

\subsection{Results}

{\bf Comparison with SOTA.} Table~\ref{tab:result2} presents a comparison of our proposed approach, DIS-Mine, against baseline models across three different datasets, including our own, ImageMine.  It is important to note that the ISD model was only evaluated on the LIS-dataset, as it requires clean, high-quality images for training, which limits its application to other datasets. The results provide a detailed analysis of two evaluation metrics: F1-score and mIoU. On the ImageMine dataset, DIS-Mine achieves the highest performance with a 70.2\% F1-score and 60.5\% mIoU, followed by the SAM model, which obtains 68.7\% F1-score and 60.0\% mIoU. For the LIS dataset, while ISD excels in low-light conditions due to its specialized training, DIS-Mine again outperforms all baselines it in the F1-score, demonstrating its adaptability to challenging environments, with ISD emerging as the second-best model. On the DsLMF+ dataset, DIS-Mine achieves the top performance with an F1-score of 86\% and mIoU of 72\%, with SAM again ranking second. These results demonstrate the robustness of DIS-Mine by achieving a strong balance between F1-score and mIoU across multiple datasets, highlighting its generalizability for diverse instance segmentation tasks.

\begin{table}[!t]
\centering
\setlength{\tabcolsep}{1.2em}  % Increase the space between columns
\renewcommand{\arraystretch}{1.5}  % Increase the space between rows
\caption{Comparing DIS-Mine vs. baselines across various datasets including our dataset, ImageMine.}
\label{tab:result2}
\begin{tabular}{p{1.8cm}|p{2.cm}|p{1.1cm}|p{1.1cm}}
\toprule
 Dataset& Model& F1-score &mIoU\\
\midrule
\midrule
\multirow{4}{3em}{\bf ImageMine \newline (Ours)} & SAM &  68.7\%  &  60.0\%\\ 

\cline{2-4}& Mask R-CNN & 65.0\%& 56.0\%\\
\cline{2-4}
& Mask2Former & 67.2\%& 58.0\%\\
\cline{2-4}
&    \cellcolor{red!20}{\bf DIS-Mine (ours)}&\cellcolor{red!20}  70.2\%  &\cellcolor{red!20} 60.5\%\\
\midrule
\midrule
\multirow{5}{5em}{LIS-dataset} & SAM &  61.0\%  &  47.5\%\\ 
\cline{2-4}
& Mask R-CNN & 58.0\%& 44.6\%\\
\cline{2-4}
& Mask2Former & 62.0\%& 45.8\%\\
\cline{2-4}
& ISD & 61.7\%& 49.8\%\\
\cline{2-4}
&    \cellcolor{red!20}{\bf DIS-Mine (ours)}&\cellcolor{red!20}  63.2\%  &\cellcolor{red!20} 47.0\%\\

\midrule
\midrule
\multirow{4}{3em}{DsLMF+} &   SAM &  84.0\%  & 71.0\%  \\ 
\cline{2-4}
& Mask R-CNN & 80.0\%& 68.0\%\\
\cline{2-4}
& Mask2Former & 83.0\%& 72.0\%\\
\cline{2-4}
&    \cellcolor{red!20}{\bf DIS-Mine (ours)}&\cellcolor{red!20} 86.0\%   &\cellcolor{red!20} 72.0\%\\
% \hline
% \multirow{4}{3em}{DarkVisionNet} &SAM    &    &   \\ 
% \cline{2-4}
% & Mask R-CNN & & \\
% \cline{2-4}
% & Mask2Former & & \\
% \cline{2-4}
% &    \cellcolor{red!20}{\bf DIS-Mine (ours)}&\cellcolor{red!20}    &\cellcolor{red!20} \\
\bottomrule
\end{tabular}
\end{table}

%In \fref{fig:learning-comparison}, we present the loss metrics for DIS-Mine and the baseline models across each epoch. This figure provides a visual comparison of how each model's training loss evolves over time, allowing us to assess the effectiveness of the training processes for DIS-Mine in relation to its counterparts. The first curve illustrates the progression of the F1-score after each epoch, providing insights into the model's classification performance over time. The second curve represents the mIoU score for each epoch, highlighting the model's ability to accurately segment instances. Lastly, the third curve depicts the training loss, which the model aims to minimize with each epoch, indicating the convergence of the training process. Collectively, these curves illustrate the effectiveness of the training regimen and the improvement of DIS-Mine across the training iterations.\red{The explanation here is not very clear. Can you rewrite this paragraph to explain which approach is the best, and what is the second good approach, and so on. Also, you can mention that one approach is better than the others in terms of F1-score while it has less performance in terms of mIoU, and so on. 

% \begin{figure*}[!ht]
%     \centering
%     \includegraphics[width=1\linewidth]{images/DIS-Mine.png}
%     \caption{DIS-Mine results on samples from our collected ImageMine dataset.}
%     \label{fig:ImageMine-result}
% \end{figure*}
\begin{figure*}[!t]
    \centering
    \subfigure[Samples from our ImageMine dataset (input images).]{
        \includegraphics[width=\linewidth]{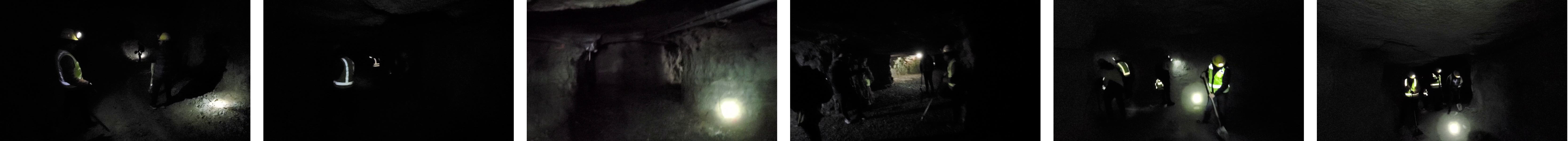}
       
    }
    \subfigure[Corresponding ground-truth masks.]{
        \includegraphics[width=\linewidth]{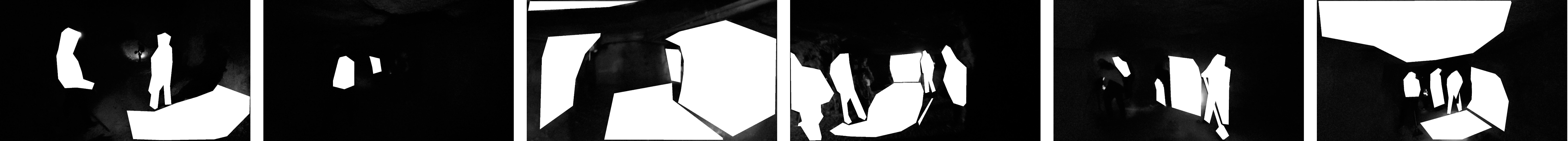}
         
    }
    \subfigure[Generated predicted masks.]{
        \includegraphics[width=\linewidth]{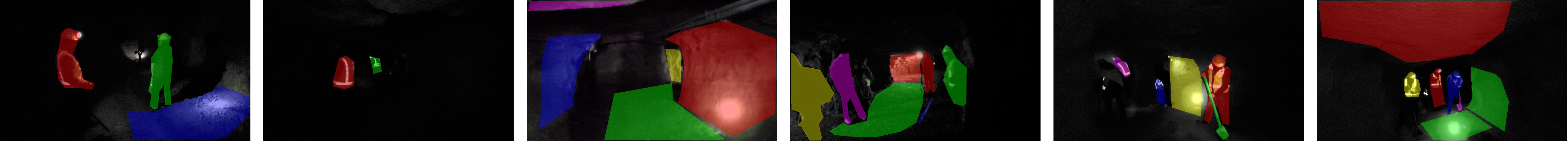}
        \label{fig:predicted-masks}
    }
    \caption{DIS-Mine prediction results on samples from our collected ImageMine dataset.}
    \label{fig:ImageMine-result}
\end{figure*}

\begin{figure*}[!t]
    \centering
    \subfigure[Samples from DsLMF+ dataset (input images).]{
        \includegraphics[width=\linewidth]{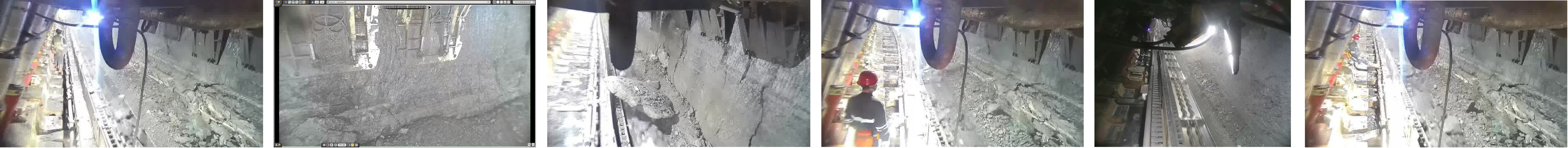}
         
    }
    \subfigure[Generated predicted masks.]{
        \includegraphics[width=\linewidth]{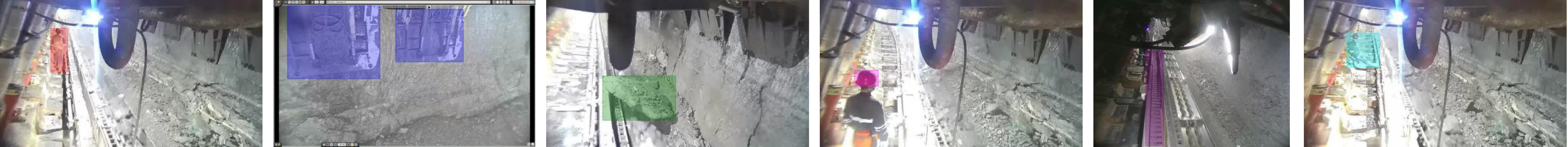}
        
    }
    \caption{DIS-Mine prediction results on samples from DsLMF+ dataset.}
    \label{fig:DslMF-result}
\end{figure*}
 
Furthermore, in \fref{fig:learning-comparison}, we show the performance trends of our DIS-Mine approach against the baselines over the training period on our ImageMine dataset, evaluating metrics like F1-Score, mIoU, and Loss.  DIS-Mine’s F1-Score curve, which plateaus at a higher value, suggests more accurate and consistent segmentation predictions. Similarly, our DIS-Mine achieves higher mIoU demonstrating its strong alignment between predictions and actual instances. The loss curves, on the other hand, reveal the models' optimization processes, where a downward trend represents error reduction and convergence toward more accurate predictions. DIS-Mine’s lower final loss value reflects better optimization, with reduced prediction errors compared to the other models.

Among the other baselines, Mask2Former consistently ranks as the second-best model in terms of mIoU and F1-Score after our DIS-Mine, indicating good segmentation accuracy but with somewhat slower convergence and less effective boundary handling compared to DIS-Mine. SAM and Mask R-CNN show relatively weaker performance, with SAM achieving moderate segmentation accuracy and Mask R-CNN displaying higher loss values and slower convergence. These limitations may stem from their challenges in optimizing boundary alignment and balancing precision with recall.

Analyzing each curve’s slope, plateau, and final values offers insight into the models’ learning dynamics. DIS-Mine stands out with its faster convergence, balanced precision-recall performance, and superior segmentation accuracy. Overall, these performance curves highlight DIS-Mine’s clear advantage in instance segmentation, demonstrating its effectiveness in both accuracy and optimization relative to the other models.

{\bf Class-Wise Performance Analysis.}
In this study, we evaluate the performance of the DIS-Mine by focusing on six object classes: people, equipment, corridor, road, wall, and roof. The study involves systematically removing one or more of these classes during training to assess the impact on overall segmentation accuracy. By excluding specific classes in different experiments, we aim to analyze the model's reliance on particular object categories and how their inclusion or exclusion affects the segmentation accuracy of the remaining classes. For example, the removal of dynamic classes such as people, equipment, and corridors enables us to examine the model's performance on static, large-scale structures like roads, walls and roofs, thereby assessing its ability to generalize without the presence of more complex or variable object classes. This methodology provides insight into the interdependence between object classes and highlights the role of class diversity in promoting the robustness and generalization capacity of the model. Our findings indicate that the model faces significant challenges when segmenting geometrically similar classes, such as roads, walls, and roofs. These classes share similar color and contrast characteristics, and due to the lack of distinctive edges, they often become indistinguishable in certain cases. This highlights the model's limitations in differentiating between visually analogous categories. As a result, we merged the road, roof, and wall classes into a single class called \textit{surrounding}. Table~\ref{tab:class_comparison} presents the performance of each class based on this merging approach.
\begin{table}[!t]
\centering
\caption{Performance based on individual class for DIS-Mine on ImageMine}
\setlength{\tabcolsep}{1.2em}  % Increase the space between columns
\renewcommand{\arraystretch}{1.5}  % Increase the space between rows
\label{tab:class_comparison}
\begin{tabular}{c|cc}
\toprule
 Class& F1-score &IoU\\
\midrule\midrule
 People&  72.6\%  & 72.6\%\\
  \hline
 Equipment & 71.4\% & 62.1\%\\
  \hline
 Corridor & 88.3\% & 78.5\%\\
  \hline
 Surrounding & 64.0\%& 52.4\%\\
\bottomrule
\end{tabular}
\end{table}

Finally, in \fref{fig:ImageMine-result} and \fref{fig:DslMF-result}, we showcase the instance segmentation outputs generated by our DIS-Mine approach on samples from the ImageMine dataset and the DsLMF+ dataset. The results highlight DIS-Mine's effectiveness in accurately segmenting instances under challenging low-light conditions. Specifically, the results in \fref{fig:ImageMine-result} demonstrate that DIS-Mine's predictions closely align with the ground truth masks for various scenarios involving people, equipment, and corridors, yielding meticulous and detailed segmentation outputs. Similarly, the results in \fref{fig:DslMF-result} show that DIS-Mine provides reliable predictions for various classes in the DsLMF+ dataset, including coal miners, hydraulic supports, large coal, miner safety helmets, towlines, and miners' behaviors.

\section{Conclusion and Future Work}
In this paper, we proposed a novel instance segmentation method named DIS-Mine to segment images of underground mines in poor-light conditions. By enhancing the images using KinD, utilizing specific prompts generated from Mask R-CNN, and aligning the masks from both the optimized SAM and Mask R-CNN models, our approach leverages the strengths of both models to achieve more accurate and robust segmentation in poor lighting conditions. The results highlight the potential of model integration and mask alignment in overcoming the limitations posed by poor light conditions because of high noise, and poor contrast. Most importantly, we collected a real-world underground mine dataset in very dark conditions called ImageMine to validate our results and compare our method with other models' performance. Our comprehensive results show that DIS-Mine outperforms each baseline model individually, delivering significant improvements in both F1-score and mIoU. Additionally, we evaluated DIS-Mine's performance on multiple low-light datasets to demonstrate the generalizability of our method across various challenging segmentation tasks.

For future work, we aim to further enhance our approach by incorporating multimodal data, such as thermal imaging and LiDAR point cloud data, to generate even more precise segmentations in complex scenarios. This integration of multimodality has the potential to significantly improve segmentation accuracy by providing complementary information, particularly in extreme low-light or adverse environmental conditions. By utilizing these diverse data sources, we hope to develop a more robust framework capable of performing in a wider range of real-world applications.

\section*{Acknowledgement}
We want to extend our gratitude to the members of the W2C lab who have participated in the data collection in the underground mine. This research is supported by a grant from CDC-NIOSH.

{\small
\bibliographystyle{IEEEtran}
\bibliography{biblio}}

\end{document}